\documentclass{article}
\usepackage{spconf,amsmath,graphicx}
\usepackage{color}
\usepackage{tabularx}
\usepackage{multirow}
\usepackage{microtype}
\usepackage{subcaption}
\usepackage{hyperref}
\usepackage{titlesec}
\titlespacing*{\subsubsection}{0pt}{0.4\baselineskip}{0.4\baselineskip}

\titlespacing*{\subsection}{0pt}{0.4\baselineskip}{0.4\baselineskip}
\titlespacing*{\section}{0pt}{0.6\baselineskip}{0.6\baselineskip}

\newcommand{\method}{Plud}

\title{Collaborative Learning of Semi-Supervised Clustering and Classification for Labeling Uncurated Data}

\name{Sara Mousavi$^{\star}$ \quad Dylan Lee$^{\star}$ \quad Tatianna Griffin$^{\dagger} $\quad Dawnie Steadman$^{\dagger}$ \quad Audris Mockus$^{\star}$}

\address{$^{\star}$  Department of Electrical Engineering and Computer Science\\
    $^{\dagger}$Department of Anthropology\\
    The University of Tennessee Knoxville, USA}
    
\begin{document}
%
\maketitle
\begin{abstract}
Domain-specific image collections present potential value in various areas of science and business but are often not curated nor have any way to readily extract relevant content. To employ contemporary supervised image analysis methods on such image data, they must first be cleaned and organized, and then manually labeled for the nomenclature employed in the specific domain, which is a time consuming and expensive endeavor.
To address this issue, we designed and implemented the \method~system.
\method~provides an iterative semi-supervised workflow to minimize the effort spent by an expert and handles realistic large collections of images. We believe it can support labeling datasets regardless of their size and type.
\method~is an iterative sequence of unsupervised clustering, human assistance, and supervised classification. With each iteration 1) the labeled dataset grows, 2) the generality of the classification method and its accuracy increases, and 3) manual effort is reduced. 
We evaluated the effectiveness of our system, by applying it on over a million images documenting human decomposition. In our experiment comparing manual labeling with labeling conducted with the support of \method, we found that it reduces the time needed to label data and produces highly accurate models for this new domain.
\end{abstract}


\section{Introduction}
\label{sec:intro}
Most published image analysis models tend to assume clean, organized image collections with ample amount of samples for training purposes. Use of these advanced models in domains that have not yet embraced 
such approaches, or for image collections that have not undergone an expansive organization, cleaning, and other type of curation, is not, unfortunately, possible. The effort we spent to overcome this problem in a collection of images documenting human decomposition led us to build the \method~system\footnote{Link to the repository: \href{https://github.com/saramsv/decaying_human_body_part_classifier}{https://github.com/saramsv/body\_part\_classifier}} to support such tasks. 

In the workflow that \method~provides, users start with a large and relatively unstructured set of images and produce highly accurate domain-specific models that can be used to clean, structure, and label the collection. That, in essence, unlocks the full potential of the content hidden in these images to solve research and practical problems. 
The main components that enables \method~to achieve its goals are: 1) the use of unsupervised clustering approaches using embeddings from an arbitrary domain to explore and design an initial grouping of images, 2) a user interface that supports rapid labeling of large batches of samples, and 3) iterations that continuously increase the accuracy of the model while reducing the labeling effort of the user. 

Specifically, the iterative sequence of clustering, labeling, and classification gradually introduce new unseen data in each iteration. The resulted labels from each iteration are used as input data for a classification model in the next iteration. High confidence results from the classifier are used to boost its learning, while low confidence decisions route images to the clustering and manual labeling component in order to be corrected. Repeating the iterations results in more labeled data and more accuracy for the classifier, and hence less need for manual labeling in consequent iterations. 

Using \method, we classified over one million images depicting the human decomposition process and evaluated the performance of our classification on a subset of 5555 randomly selected and manually labeled images. The results show we were able to classify these images with top-1 and top-3 F-score of 79.89 and 93.84  respectively.

Using unsupervised methods one can cluster image data using their numerical embeddings \cite{guerin2017cnn, yang2016joint, mousavi2019analytical, wang2017unsupervised} in groups that share similar characteristics. However, there is no certainty that such methods will be able to recognize any relevant classes for a new domain without manual evaluation, which is time consuming. 


Other works on semi-supervised and weakly supervised learning methods 
\cite{dopido2013semisupervised, rosenberg2005semi, yarowsky1995unsupervised,lee2017deep, grimaudo2014select} target speeding up the data labeling process and other computer vision tasks. These approaches start with a small set of labeled data and a large set of unlabeled data. The labeled data then guides the learning process to make the models more generalizable to the remaining unlabeled data. 
In these works, transfer learning and fine-tuning are often used to extend what is learned from one dataset to other datasets \cite{raji2017photo, mahajan2018exploring}. Unfortunately, labeled data that can represent the new dataset is needed for transfer learning and fine-tuning to perform well. This is especially the case when the new dataset has a completely different set of classes.


Ways to find samples that are hard to learn was considered in~\cite{xue2019hard}, and we incorporate similar techniques in \method.
A workflow for labeling by Philbrick et al.~\cite{philbrick2019ril,ma2019cnn} supports annotating medical images involving an iterative approach of manual labeling and classification. However, their system works well only for small medical datasets. 


The rest of the paper is as follows. Detailed information about our proposed system is provided in Section \ref{sec:method}. Section \ref{sec:resdis} includes an introduction to the dataset and the conducted experiments and their results. We finally conclude the paper in Section \ref{sec:con}.

\section{Method}
    \label{sec:method}
\begin{figure}
        \centering
        \includegraphics[width=.45\textwidth]{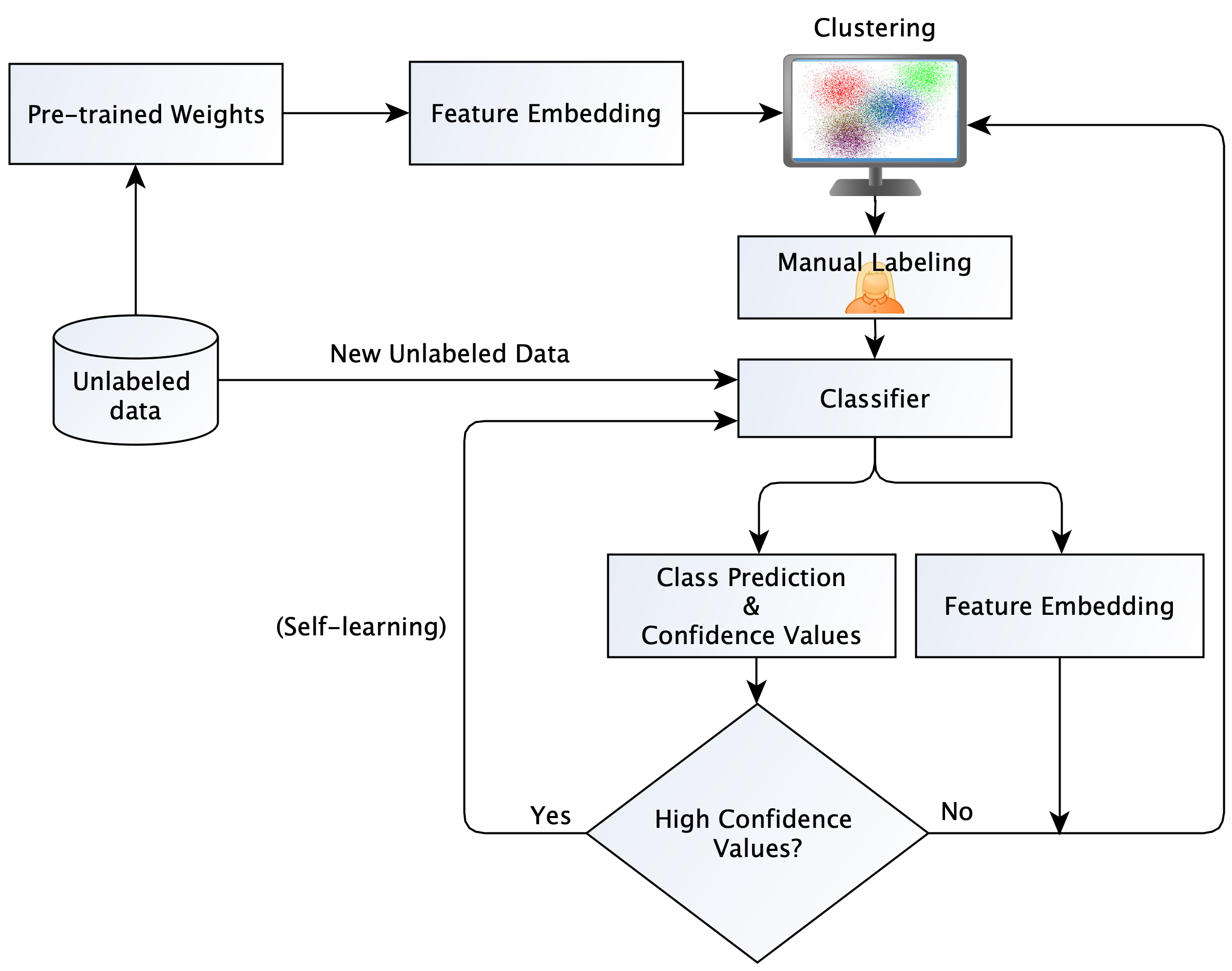}
        \caption{Shows the overall architecture of \method. Unlabeled data are mapped to numerical feature embeddings and then clustered together. The resulted clusters are displayed to the domain expert for manual labeling through the \method~interface. The labeled data is used for training a classifier by which, for a new set of unlabeled data, labels and feature embeddings are  generated. The iteration repeats for images with low confidence predictions. High confidence predictions are fed back to the classifier to enable self-learning. }
        \label{fig:arc}
    \end{figure}

    Figure \ref{fig:arc} shows an overview of \method. The basic idea of \method~is to combine unsupervised clustering, and supervised classification methods in an iterative workflow involving a human expert, with a user interface tailored for enabling rapid labeling of a large number of images. In each iteration, the uncertainty of the classifier is used to  suggest additional data to be clustered and then manually labeled. Images that are considered similar are presented in large batches to leverage the power of the human visual system to detect outliers. The domain expert produces names/labels for each group based on the dominant class in the cluster. The resulting labeled data are then used for training and fine-tuning the classifier in the next iteration. 
    
    In the following, details about data preparation, clustering, the labeling interface, and the classification step are provided.
    
\subsection{Data Preparation}
    \label{sec:sample}
    The performance and generalizability of classifiers depends on the quality and the size of their training data. The more diverse and representative the training data is, the more generalizable the models are likely to be. In the case of a temporal dataset, such as images documenting human decomposition in which the subjects' appearance changes over time, it is important to sample the training data in such a way that includes the dataset's characteristic. In our case, for example, images of all possible decomposition phases should be included in the training data. To ensure the inclusion of changes over time on the human bodies, we randomly selected a small number of subjects and then selected all images taken from them over time, from fresh to decay, rather than randomly selecting images from the entire set which might include more subjects but might also exclude some decomposition phases.

\subsection{Clustering} \label{sec:cluster}
    In order to assist and speed up the manual labeling process and enable mass labeling, we built a web-based interface that clusters images so that they can be viewed and evaluated by users.
    To build the clusters, the unlabeled data are mapped to feature embeddings using a convolutional neural network pre-trained on Imagenet~\cite{deng2009imagenet}. In the first iteration that we had no labeled data, we used a ResNet-based model. We fed in the images to the network and extracted the output of the convolutional layers as the feature embeddings and then clustered them. For subsequent iterations, we used the feature embeddings generated by our trained classifier model. We used various clustering methods such as agglomerative clustering~\cite{mullner2011modern}, spectral clustering~\cite{stella2003multiclass} and KMeans~\cite{krishna1999genetic} to group similar images together. From a manual examination of the clusters we concluded that KMeans outperformed the rest of the methods for our dataset.
    
    The resulted clusters are displayed to a user for evaluation and cleaning through a web interface. The interface allows the user to see all images assigned to one cluster in a single web page. Each image can be selected by a single click if they are misclustered. The rest of the cluster is labeled with the label that matched the majority of the images in the cluster. 
    
\subsection{Classification}
We use the labeled data resulting from the clustering-based approach to train a classifier model. The output of our classifier for each input is a feature embedding which is resulted from the convolutional layers only, a label and a confidence value for the given label. 
The initial training data for the classifier is obtained from the labels given by the expert to the  initial clusters. During subsequent iterations, the classifier is used to make predictions on the batches of unlabeled data, and the predictions are ordered by the confidence level the classifier assigns to each prediction. High-confidence predictions are used to expand the training data using the predicted labels and 
low confidence predictions are assigned for clustering and manual labeling. The threshold for considering high or low confidence values is determined by manually exploring the predictions. 

Adding images with high confidence predicted labels to the classifier's training data improve its accuracy on those type of images. Manual labeling of the images with low confidence predictions is still necessary to accurately predict the class for the instances where the certainty of the classifier was low. 
In other words, the low confidence values indicate the types of data the classifier has not yet fully learned. To expose the classifier to such data, we use the feature embedding outputted from the classifier for these images to cluster them. The clusters are then displayed to a domain expert for labeling and verification. Note that these embeddings are better representations of the images than those from the initial Imagenet-trained model as they have been partially trained on the data from the domain. 

\section{Evaluation}\label{sec:resdis}
We have tested our method on a dataset of images capturing the process of human decomposition. In the following section we provide more details about the dataset, the conducted experiments and the end results. 
\subsection{Dataset} \label{sec:ds}
Images tracking human decomposition process are of great value for conducting research. This is due to two main reasons. First, when using images instead of actual decaying bodies, studying the human decomposition process is no longer limited to only the period of decomposition itself.
Second, it allows research to focus on particular factors, such as mold formation or mummification, and correlate such factors to both environmental and individual characteristics (e.g., age or weight).

The image collection consists of photos taken of decomposing humans donated to the Forensic Anthropology Center of our university. These donors are placed at the Anthropology Research Facility ("Body Farm") where the different stages of decomposition are studied and recorded through photography. An identity number is assigned to each donor and written on a wooden stake and placed next to them. The pictures are taken periodically from various angles of the bodies and capture different body parts to show the stage of body decomposition at any given time. The collection has over one million images and images from following categories:  \textit{Arm}, \textit{Hand}, \textit{Foot}, \textit{Legs}, \textit{Full Body}, \textit{Head}, \textit{Backside}, \textit{Torso}, \textit{Stake}, \textit{Plastic}.

For this work, we selected and labeled $5555$ images capturing all images taken from $3$ subjects as our ground truth data for testing \method. The performance metrics (average precision, average recall and F-score) reported in this paper are resulted using this test data.

\begin{figure*}[]
    \includegraphics[width=.30\textwidth]{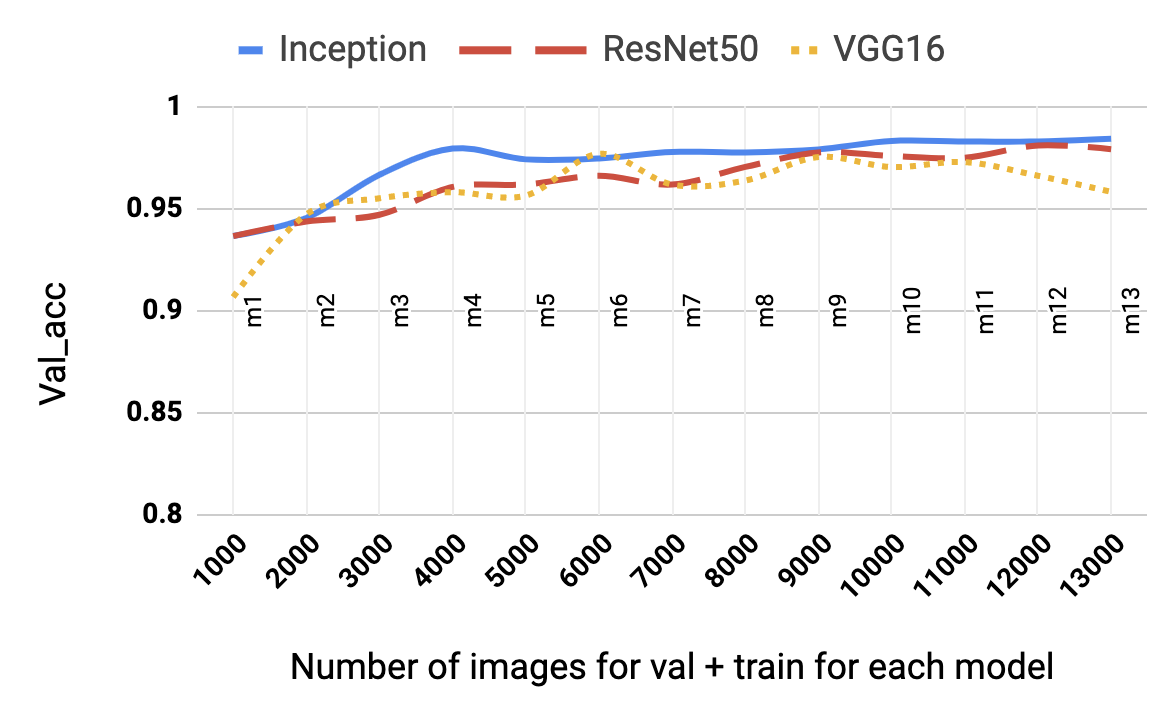}\hfill
    \includegraphics[width=.30\textwidth]{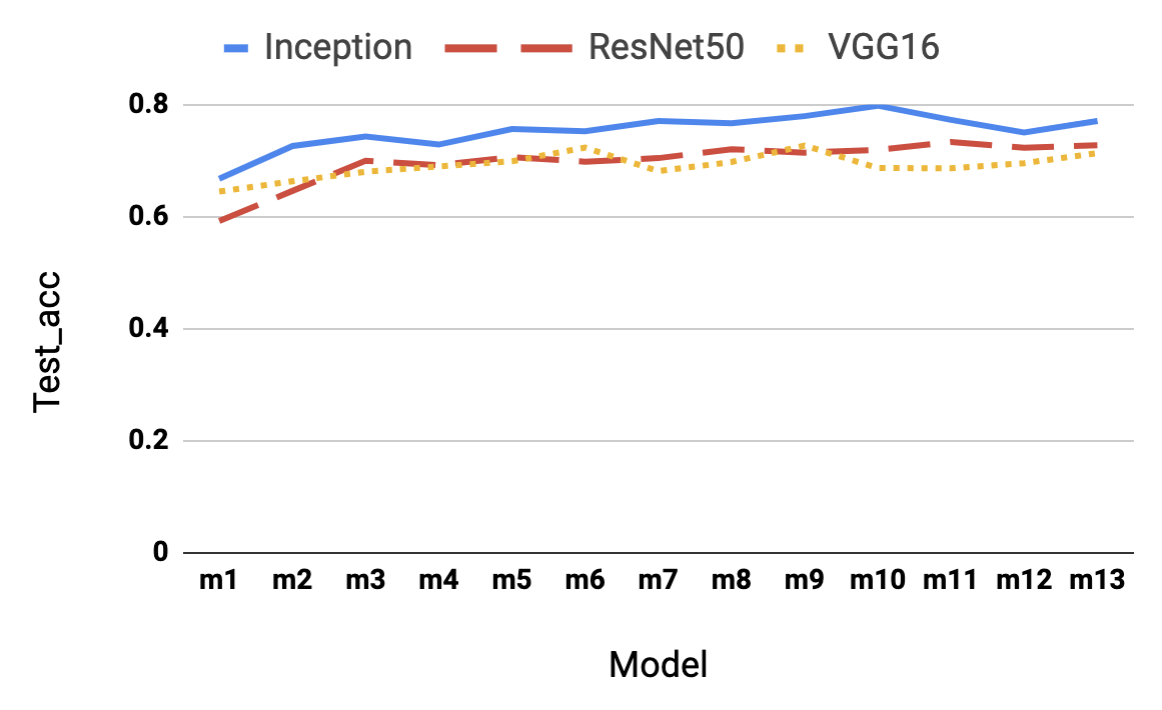}\hfill
    \includegraphics[width=.30\textwidth]{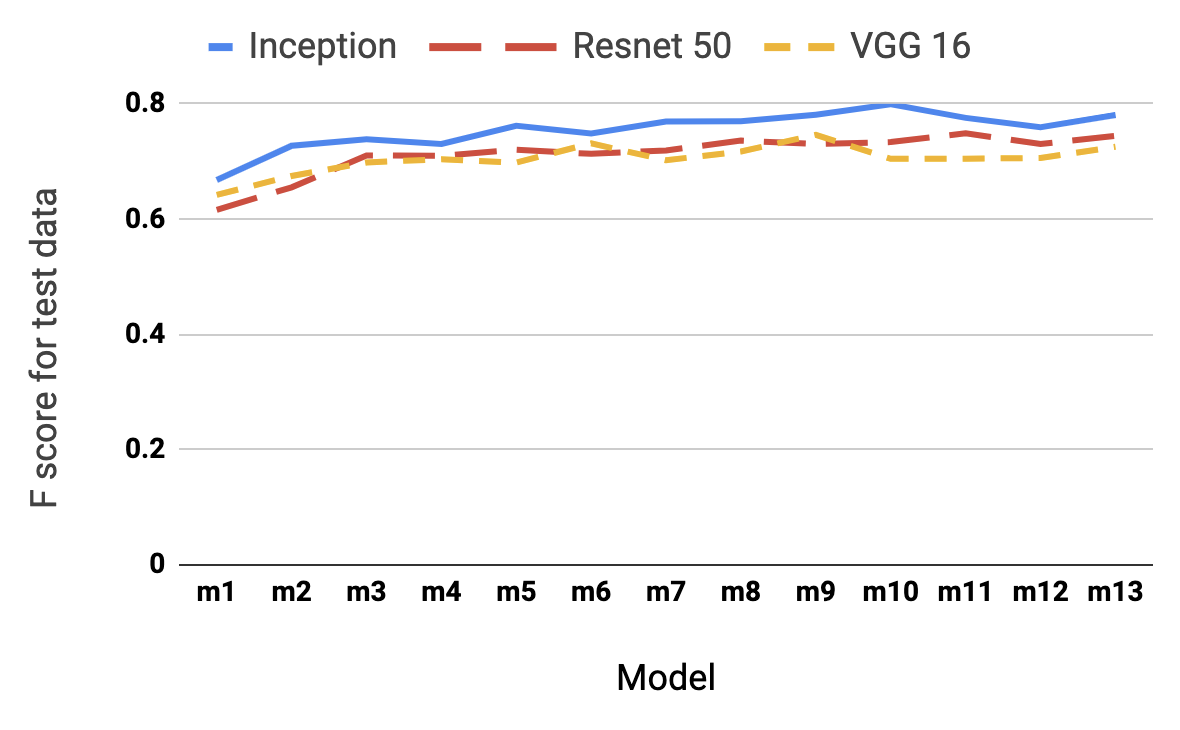}\hfill
    \caption{The performance of models trained on various number of images is shown. Test\_acc and F score are calculated using the manually labeled test data mentioned in Section \ref{sec:ds}. $\left \{ m_1, m_2, \cdots, m_{13} \right \}$ are the models that are refined with more data through $13$ iterations. Although it is expected to have better performance as the amount of training data increases, it is important to note that, the more accuracy in each iteration results in less effort for the user in the labeling process as shown in Table \ref{tbl:time}.}
    \label{fig:size}
\end{figure*}
\begin{table*}[]
\begin{tabular}{|c|c|c|c|c|c|c|c|c|c|c|c|c|}
\hline
\multirow{2}{*}{\textbf{Model}}      & \multirow{2}{*}{\textbf{}} & \multicolumn{10}{c|}{\textbf{Precision of Classes}}                                                                                                                        & \multirow{2}{*}{\textbf{AP}} \\ \cline{3-12}
&     & \textbf{Arm} & \textbf{Hand} & \textbf{Foot} & \textbf{Legs} & \textbf{Full Body} & \textbf{Head} & \textbf{Backside} & \textbf{Torso} & \textbf{Stake} & \textbf{Plastic} &\\ \hline
\multirow{2}{*}{\begin{tabular}[c]{@{}c@{}}Inception\end{tabular}} & Top 1                 & 45.73 & 85.60 & 93.72 & 60.52 & 92.69 & 94.33 & 68.25 & 87.22 & 96.30 & 73.61 & \textbf{79.80} \\ \cline{2-13}                                                                          
& Top 3 & 80.23 & 96.86 & 97.75 & 86.96  & 97.53 & 98.29 & 89.57  & 97.20 & 98.87 & 88.88 & \textbf{93.21}\\ \hline
\hline
\multirow{2}{*}{\textbf{Model}}  & \multirow{2}{*}{\textbf{}} & \multicolumn{10}{c|}{\textbf{Recall of Classes}}                                                                            &\multirow{2}{*}{\textbf{AR}} \\ \cline{3-12}
&  & \textbf{Arm} & \textbf{Hand} & \textbf{Foot} & \textbf{Legs} & \textbf{Full Body} & \textbf{Head} & \textbf{Backside} & \textbf{Torso} & \textbf{Stake} & \textbf{Plastic} &                              \\ \hline
\multirow{2}{*}{\begin{tabular}[c]{@{}c@{}}Inception \end{tabular}} 
& Top 1  & 53.88 & 67.36 & 62.34 & 97.60 & 77.21 & 94.91 & 55.84 & 91.97 & 98.86 &1 & \textbf{80.00} \\ \cline{2-13} 
& Top 3 &92.69 & 87.63 & 90.66 & 99.57 & 96.34 & 99.75 & 81.81 & 96.27 &1.   &1. & \textbf{94.47}\\ \hline
\end{tabular}
\caption{Shows precision and recall for our Inception-based classifier on the test data.}
\label{tbl:PR}
\end{table*}
\subsection{Results}
In this section, we evaluate the efficiency gained from our labeling interface in addition to the affects of data sampling and the size of training data on the labeling performance. 

\subsubsection{Labeling Interface Performance}
In order to examine how our clustering-based web interface affects the speed of labeling, we randomly selected $150$ images and asked a user to label them with and without using our interface. We then repeated the same process for 300 images to see if the labeling time linearly increases when we double the number of images. 
In order to have a fair comparison, we designed a similar interface for the user with and without clusters. For the basic case that does not involve clustering, image thumbnails are generated on a single web page where the user can select the desired images. Using this interface, the user selects images belonging to the same class in each iteration until all images are labeled with appropriate classes. In the cluster-based interface of \method, images are displayed based on the cluster they belong to. Thus the user only needs to select the miss-clustered images and label the cluster based on the majority of the images within it. 
The result of this experiment is shown in Table \ref{tbl:time}. The result shows that our interface drastically reduces the labeling time. It also shows that a linear increase in the number of images non-linearly increases the time of labeling. The result also indicates that labeling smaller batches of data is faster. We believe that this is due to the visual confusion caused by scrolling through large number of images. 

\begin{table}[]
\begin{tabular}{|c|c|c|c|}
    \hline
    & 150 images & 300 images\\ \hline
   Basic labeling &   13m.22s   &  43m.52s \\ \hline
   Unsupervised assisted labeling &    4m.41s &  11m.34s \\ \hline
\end{tabular}
\caption{In the Basic method, the user sees all images at once and labels them. In the second approach, the images are first grouped using our clustering interface and the user labels the groups and corrects the misclustered images. The time for 300 images shows that the increase in the number of images non-linearly adds to the labeling time.}
\label{tbl:time}
\end{table}

\subsubsection{Data Sampling Strategy}
As mentioned in Section \ref{sec:sample}, it is important to make sure that the train data is diverse and represents the structure of the dataset. 
We designed an experiment to test this hypothesis by having a set of $5000$ randomly selected images from the entire dataset and $5000$ images from multiple randomly selected subjects while making sure it includes all stages of decomposition. We trained our classifier on the two sets and tested it on our test set. The results shown in Table \ref{tbl:decom_ds} indicate that the classifier trained on the structure-aware multi-subject data performs with $76.71\%$ accuracy on the test set where as the accuracy is $74.51\%$ for the classifier trained on randomly selected data.

\subsubsection{Classifier Performance}
We analyzed the effect of adding data in each iteration on the performance of the classifier to ensure that in each iteration we have a more accurate model and thus less effort on the domain expert for manual labeling. We used three different models to find the model with highest accuracy. We checked the performance of VGG16, ResNet50 and Inception-based classifiers trained on various amounts of data, $\left \{ 1000, 2000, \cdots , 13000 \right \}$. In each iteration, a new set of $1000$ unlabeled images were fed to the classifier. High confidence predictions were used as new labels and images with low confidence predictions were labeled through the clustering interface. The new $1000$ labeled images were added to the training data. 

When the classifier is trained, its performance on a test set of $5555$ manually labeled images was examined. The vall\_acc, test\_acc and F-score are shown in Figure \ref{fig:size}. Increasing the size of the training data in all three classifiers resulted in an overall increase in their performance on our test data. It is important to note that although this result is expected, in the context of \method~that means less manual effort on labeling for the domain expert since less misclustered images need to be manually selected each time. 

Based on the results shown in Figure \ref{fig:size}, we obtained the best accuracy when using an Inception-based classifier which is then used to calculate precision and recall for test data shown in Table \ref{tbl:PR}. 

\section{Conclusion}\label{sec:con}
We proposed \method~as a system for speeding up the labeling of uncurated large collections of images in cases where no labeled data pre-exists and the types of images, and relevant image classes are difficult to assess or are not available in advance. For evaluation, we used \method~to label images depicting human decomposition. Our evaluation shows that \method~accelerates and facilitates labeling of large amounts of data. We were able to easily label $13192$ and $5555$ train and test images respectively using the iteration provided in \method. Our trained classifier on these labeled data was then used to automatically label the remaining $1,148,511$ images.


 \begin{table}[]
\begin{tabular}{|c|c|}
    \hline
    & accuracy\\ \hline
    Random sampling & 74.51 \\ \hline
    Structure-aware multi random subject sampling &   76.71 \\\hline
\end{tabular}
\caption{Shows the performance of an Inception based classifier when trained on randomly selected images vs random multi-subjects' data.}
\label{tbl:decom_ds}
\end{table}



\clearpage
\bibliographystyle{IEEEbib}
\bibliography{refs}

\end{document}